\documentclass[sigconf]{acmart}

\usepackage{booktabs} 
\usepackage{amsmath}
\usepackage{mathtools}
\usepackage[super]{nth}
\usepackage{listings}
\usepackage{amssymb}
\usepackage{pgfplots}
\usepackage{url}
\usepackage[]{appendix}
\usepackage{caption}

\setcopyright{rightsretained}

\acmConference[MILETS17]{3rd SIGKDD Workshop on Mining and Learning from Time Series}{August 2017}{Halfiax, Nova Scotia Canada} 
\acmYear{2017}
\copyrightyear{2017}

\begin{document}
\title{Robust Parameter-Free Season Length Detection in Time Series}

\author{Maximilian Toller}
\affiliation{%
  \institution{Institute of Interactive Systems and Data Science}
  \institution{Graz University of Technology}
  \streetaddress{Inffeldgasse 13}
  \city{Graz} 
  \state{Austria} 
  \postcode{8010}
}
\email{maximilian.toller@student.tugraz.at}

\author{Roman Kern}
\affiliation{%
  \institution{Institute of Interactive Systems and Data Science}
  \institution{Graz University of Technology}
  \institution{Know-Center}
  \orcid{0003-0202-6100}
  \streetaddress{Inffeldgasse 13}
  \city{Graz} 
  \state{Austria} 
  \postcode{8010}
}
\email{rkern@know-center.at}

\renewcommand{\shortauthors}{Toller, Kern}

\begin{abstract}

The in-depth analysis of time series has gained a lot of research interest in recent years, with the identification of periodic patterns being one important aspect. 
Many of the methods for identifying periodic patterns require time series' season length as input parameter. 
There exist only a few algorithms for automatic season length approximation.
Many of these rely on simplifications such as data discretization and user 
defined parameters. 
This paper presents an algorithm for season length detection that is designed 
to be sufficiently reliable to be used in practical applications and does not
require any input other than the time series to be analyzed. 
The algorithm estimates a time series' season length by interpolating, 
filtering and detrending the data.
This is followed by analyzing the distances between zeros in the directly 
corresponding autocorrelation function. 
Our algorithm was tested against a comparable algorithm and outperformed it by passing 122 out of 165 tests, while the existing algorithm passed 83 tests. 
The robustness of our method can be jointly attributed to both the algorithmic approach and also to design decisions taken at the implementational level. 


\end{abstract}

%
%



\maketitle
\newcommand{\bff}{\textbf}
\newcommand{\mbeq}{\overset{!}{=}}
\newcommand{\mnbeq}{\overset{!}{\neq}}
\newcommand{\addtag}{\refstepcounter{equation}\tag{\theequation}}

\section{Introduction}


In many areas of natural science and economics there exists an abundance data 
which can be used beneficially once they are processed. Often these data are 
collected at regular intervals to enable making statistical assumptions and
predictions for numerous applications. 
Data collected repeatedly over a time span is referred to as \textit{time series}.

The in-depth analysis of time series has been a central topic of research in 
recent years. Typically, time series are investigated to better understand their past
behavior and to make forecasts for future data. To achieve this, identifying
trends and regularities are particularly useful, as they allow one to generalize
from the given data to a larger context.

A very natural approach to finding these relevant patterns is the Exploratory
Data Analysis~\cite{Tukey}. This method relies on visualizing the data, typically
without making any prior assumptions about the data. An advantage of this 
procedure is that humans usually analyze visual data very effectively and thus
identify many relevant features of the data.
\begin{figure}[!h]
\centering
\includegraphics[scale=0.45]{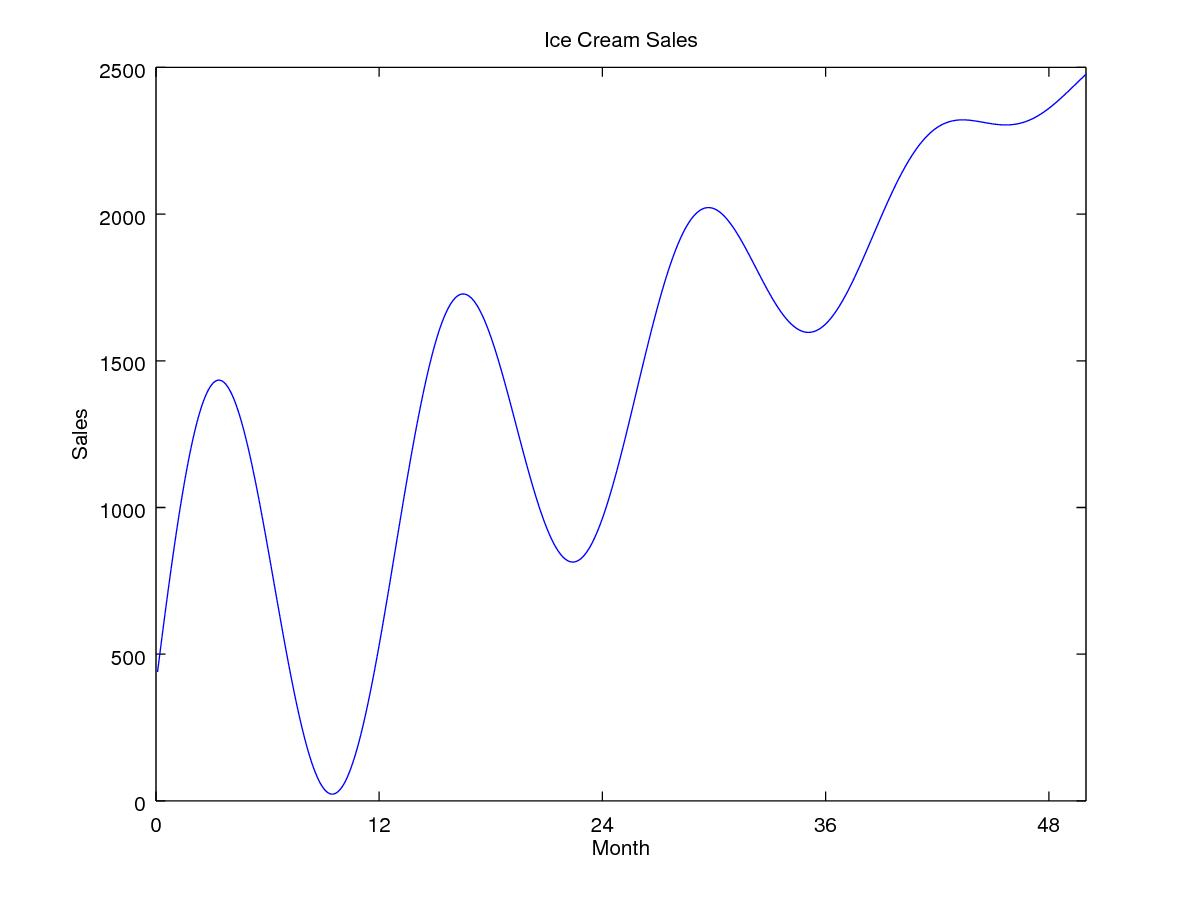}	
\caption{A simple example of a time series with trend and periodic patterns. 
Here the season length is 12 month.}
\label{fig:example}
\end{figure}

While data visualization works well with many time series, there are also
several cases where it is insufficient. When dealing with very complex data,
it is often difficult to infer meaningful features. Furthermore,
it is also frequently necessary to analyze dozens of time series, where manual
analysis quickly becomes tedious or even infeasible. 
Therefore, an automated and robust approach that can deal with complex data is required.


While there are many established algorithms for trend estimation~\cite{Draper}~\cite{Vamos}, 
methods for 
finding periodic patterns and features are much younger. One popular approach is 
the data cube concept~\cite{Han1}, which relies on constructing and employing
a data cube for finding periodic features in a time series. This method would, 
for instance, successfully identify a peak of ice cream sales every summer given
monthly sales data over several years. However, to achieve this, the algorithm
requires the user to input the time series' \textit{season length} (or \textit{frequency}). In
the example depicted in Figure~\ref{fig:example} the season length would be 12, as the sales pattern repeats
every 12 months.

%
%

Today, there already exist a few algorithms for approximating season length. 
Most of them rely on data discretization before analysis, which is
a simplification of data as seen above to discrete numbers such as 2, 4, and 6.
For finding so-called \textit{symbol periodicity} in discretized data one can
for instance use suffix trees~\cite{Rasheed1} or convolution~\cite{Elfeky1}. 
One downside of discretization is that it inevitably leads to a loss of 
information, which might change the result in some cases. A more 
significant disadvantage of discretization is that contemporary time series symbolization methods such as SAX~\cite{Lin1} require the user to define
the number of symbols to which the data is discretized. This is be problematic
since data mining algorithms should in general have as few parameters as 
possible~\cite{Keogh1}.

A different approach that does not rely on data discretization is searching
for local peaks and troughs in a time series' autocorrelation function
\cite{Hyndman1}. This method works fully automated and efficiently computes
the correct season length of many time series. 
Yet there are still several cases where this algorithm is not able to identify the correct frequencies - particularly in time series with noisy data or very long periods.

Due to the above mentioned algorithm's disadvantages, it should be beneficial to
find a more reliable method. Such an algorithm should not rely on user defined 
parameters or 
data discretization and still correctly identify a time series' season length.
A method that was designed to meets these requirements is presented in this paper.
The source code  of our system, the test data set and detailed test results can be accessed online\footnote{\url{https://github.com/mtoller/autocorr_season_length_detection}}.

\section{Background}

The existing time series analysis literature has proposed several different approaches for seasonality detection. 
Often the term periodicity detection is used in literature for the same task.
These methods can be split in three different categories: explicit season length, discretized time series, and single season length.

\paragraph{1. Explicit season length}
The first type of algorithms depends on an external specification of the period
length to extract periodic features. The previously mentioned data cube
concept~\cite{Han1} was a pioneer method for mining periodic features in
time series databases. Other examples are the chi-squared 
test~\cite{Ma1} or binary vectors~\cite{Berberidis1}. The
advantage of such procedures is their low time complexity and their reliable 
results. However, relying on an external specification of the period inherently prevents these methods from detecting an unknown period. 
In many applications a parameter-free method is being highly desired.

\paragraph{2. Discretized time series}
The second type of algorithms for detecting periodic features test all possible
periods of a time series. This is typically performed on a discretized time 
series, as otherwise such algorithms would quickly become computationally 
infeasible. Examples for such methods are suffix trees~\cite{Rasheed1}, convolution~\cite{Elfeky1} and data sketches~\cite{Indyk1}. Unlike
the previous type, these algorithms typically find all potentially relevant 
periodic patterns without previous specification of the period. However,
since they depend on data discretization, they can only be as
accurate as the underlying data discretization allows. Furthermore they may return more than 
one possible result. While this is often an accurate reflection of a time series
periodic features, it is often advantageous in practice to provide a single
dominant period for further automated processing of a time series. 

\paragraph{3. Single season length}
The third type of algorithms approximates a single season length from the raw data. 
While such an approach might be too limited for highly complex data, it can be useful
in cases where time series have one dominant frequency. One popular algorithm~\cite{Hyndman2} of this category can be found in the R forecast 
library and was derived from another method~\cite{Hyndman1} which is based on 
autocorrelation.
The algorithm presented in this paper belongs to the third type and focuses on improving 
the robustness of the season length detection.

\paragraph{Combination of methods}
All three categories consequently serve different purposes in time series
seasonality detection and also can be combined in useful ways. For 
example, an algorithm of type 3 can be used for season length detection, which
can then be used as input for a type 1 algorithm to enable correct periodicity 
mining. Further, type 2 and 3 can be combined on the same time series to assess
the effect discretization has on a specific time series. 


\section{Concepts}
Let $x = \{x_1,x_2,...,x_n\}$ be a series of real-valued observations and $\Delta_x$ the
interval at which these observations are made. If $x$ has a subsequence of 
observations $Q_x$ which
occurs every $s$ observations after its first occurrence, $x$ is seasonal. 
Moreover, if $Q_x$ is not part of a longer repeating
subsequence of $x$ and it cannot be divided into shorter equal subsequences, then
$x$ is perfectly seasonal with season length $s\Delta_x$.

$\Delta_x$ is negligible since 
it is independent from $x$. The problem of finding season length $s\Delta_x$ can 
therefore be reduced to finding the characteristic subsequence $Q_x$ which fulfills the following formal 
criteria~\ref{eq:criterion1} and \ref{eq:criterion2}. The first criterion is
\begin{equation}
	|Q_x| = s
	\label{eq:criterion1}
\end{equation}

 which means the number of observations in $Q_x$ must be 
equal to the number of observations in $x$ from one occurrence of $Q_x$ to its 
next. The second criterion is 
\begin{equation}
\forall q \in \rho(Q_x)\setminus Q_x : q^k \neq Q_x
\label{eq:criterion2}
\end{equation}
 with
$\rho(Q_x)$ being the power set (set of all subsets) of $Q_x$ and $q^{k} :=
q^\frown q^{k-1}$ ($q$ followed by $q^{k-1}$). This means that $Q_x$ must not contain a subsequence q
that can be repeated $k$ times to create $Q_x$, i.e. $Q_x \neq \{q,q,q,...\}.$

For example, given a time series 
\begin{equation}
 y = \{0,2,1,2,0,2,1,2,0,2,1,2,0,2,1,2\}
 \label{eq:example}
\end{equation}
the sequence $a = \{0,2,1,2\}$ is repeated four times in $y$.
Sequence $a$ cannot be reduced to shorter equal sequences, as $a_{1,2} = \{0,2\} \neq
a_{3,4}=\{1,2\}$ and the number of observations in $a$ is equal to the number 
of observations from one occurrence of $a$ to its next occurrence $|a| = s_a = 4$.
Therefore $a = Q_y$ and $y$ has a season length of $4\Delta_y$.
 
If one extends the sequence by one element to $b=\{0,2,1,2,0\}$ then $|b| = 5 
\neq s_b = 8$, which means it cannot be $Q_y$. Now considering
sequence $c = \{0,2,1,2,0,2,1,2\}$ the first criterion is fulfilled: $|c| = s_c  = 8$. 
Yet the second condition is violated as $c$ can be split into the equal 
subsequences 
$c_{1,2,3,4} = c_{5,6,7,8} = \{0,2,1,2\}$, therefore $c$ can be divided into 
shorter equal sequences and also cannot be $Q_y$.

\subsection{Detrending}

Many time series have a trend in addition to its seasonality, which means that
the seasonal influences revolve around a trend function. For instance, in the
example in Equation~\ref{eq:example} there could be a linear increase in the observations after every 
season, e.g. 
\[y'=\{0,2,1,2,0.1,2.1,1.1,2.1,0.2,2.2,1.2,2.2,...\}\]
In this case $Q_{y'}$ would be time-dependent: 
\[Q_{y'}(t) = \{0,2,1,2\} + \frac{1}{10}*\lfloor\frac{t}{4}\rfloor\]
with $t \in \mathbb{N}_0$.
However, the unscaled season length $s_{y'} = |Q_{y'}|$ is identical for all $t$ 
and is therefore not time-dependent. Hence it is desirable to remove trend 
influences from a time series before analyzing seasonality.

Approximating the trend components can be achieved with regression analysis~\cite{Draper}, which
is a procedure for finding a function which minimizes the mean square error 
between the observations and the approximated function. Its cost function can 
be written as
\begin{equation}
C(\theta) = \frac{1}{N} \sum_{i=1}^{N}((X_i)^T \theta - x_i)^2 
\label{eq:mse}
\end{equation}
where $X$ is the design matrix, $x$ the time series and $\theta = 
\{\theta_1,...\theta_n\}$ the parameters of the regression.
A design matrix is a matrix where each row describes one observation
and each column models an assumed feature onto the corresponding observation.
For instance, in polynomial regression $X$ can be written as
\[
X =
\begin{bmatrix}
1 & 1 & 1 & 1 & ...\\
1 & 2 & 4 & 8 & ...\\
1 & 3 & 9 & 27 & ...\\
1 & 4 & 16 & 64 & ...\\
\vdots & \vdots & \vdots & \vdots& \ddots
\end{bmatrix}
\]

To find the parameters which minimize the cost function, one can use the 
analytical solution

\begin{equation}
	\theta = (X^T X)^{-1} X^T x.
	\label{eq:linreg}
\end{equation}

The trend  of time series $x$ can then be removed with 
\begin{equation}
	x_{detrend} = x - X	\theta.
	\label{eq:detrend}
\end{equation}

\begin{figure}[!h]
	\centering
	\includegraphics[scale=0.45]{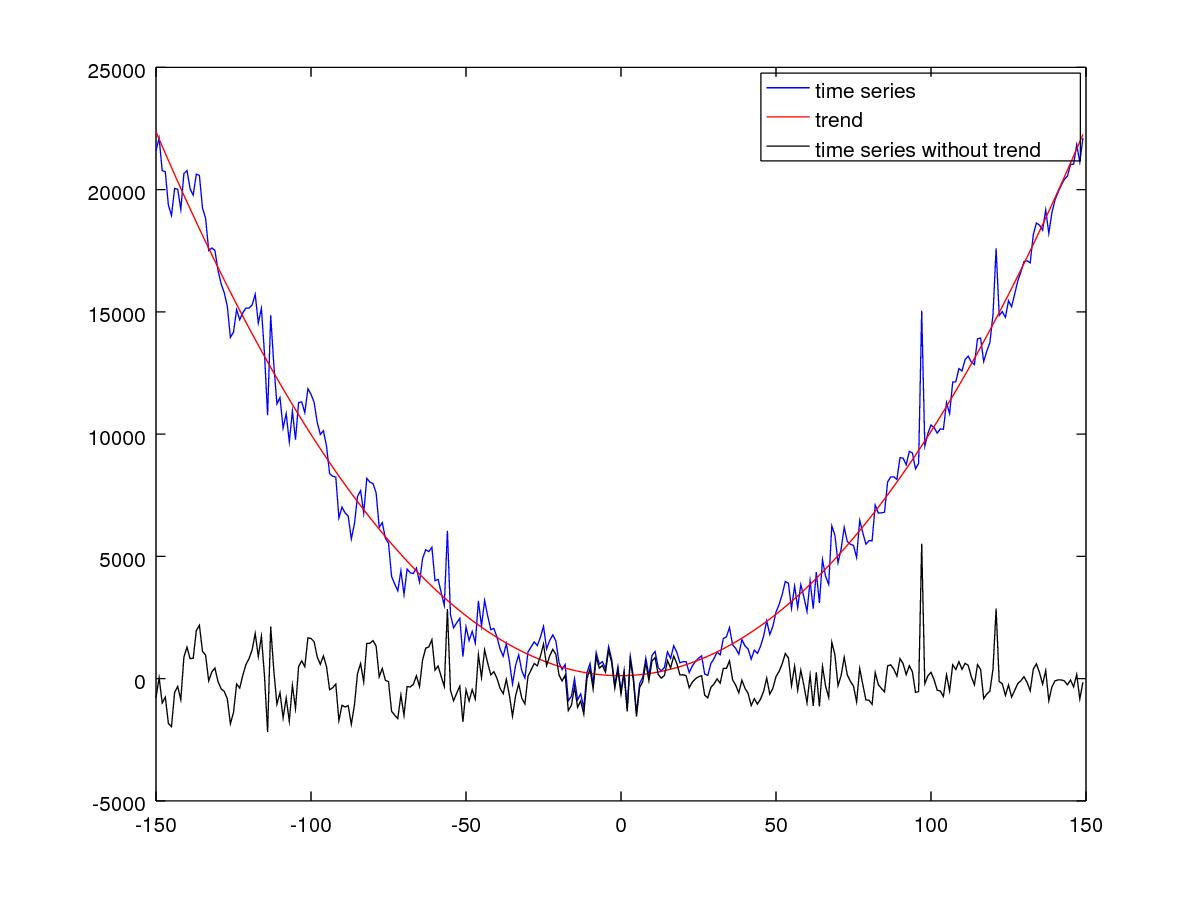}	
	\caption{Examples of time series with (blue) and without (black) quadratic trend.}
\end{figure}

Removing all trend influences from a time series significantly facilitates 
investigating seasonal effects, as the characteristic subsequences $Q_x^{(t)}$ 
then directly correlate with each other. This property allows one to investigate
the time series' autocorrelation instead of the time series itself. Such an
approach is advantageous since the original observations are frequently much 
more difficult to analyze than autocorrelation.

\subsection{Removing Noise}

Before directly analysing the characteristic subsequence $Q_x$, another major component of many time series needs to be considered. 
Most non-discretized time series contain random inaccuracies, which are often named noise or residuals. 
Minimizing these random influences is desirable, since they tend to obscure the true nature of the 
observations. This is typically achieved by means of filters.

As noise typically affects every observation in a time series, it tends to
have a shorter period than seasonality, which has a period of at least 2 
observations. Such a shorter period leads to noise having a higher frequency 
than seasonality. Therefore, it is advantageous to apply a low-pass filter on 
the time series, which rejects any frequency higher than a given threshold. This
smooths the curve while maintaining the original season length.

However, using a low-pass filter also has a negative effect on the 
observations. While removing white noise, filters likewise alter the
observations' correlation. 
To lessen these unwanted side effects, one may interpolate the data before applying the low-pass filter.

\subsection{Analyzing Correlation}
Only after removing most noise from the observations, it is possible to meaningfully investigate autocorrelation of a time series.
A time series' autocorrelation is the correlation of its observations at different times. 
Autocorrelation is formally defined as
\begin{equation}
A_x(\tau) = \int_{-\infty}^{\infty}x_t\overline{x}_{t+\tau}dt
\label{eq:acorf1}	
\end{equation}

where $\tau$ is the lag at which the observations are compared, and 
$\overline{x}$ is the complex conjugate of $x$. Since time series are 
time-discrete and real-valued, this can be further simplified to

\begin{equation}
A_x(\tau) = \sum_{t}^{T}x_t x_{t+\tau} .
\label{eq:acorf2}	
\end{equation}

After a further normalization the resulting values range from $-1$ to $+1$. 
The former implies complete anti-correlation between the two observations, 
while the latter means full correlation.

In general, two objects $a$ and $b$ correlate, if a change in
$a$ is likely to be linked with the same change in $b$ as well, whereas negative 
correlation would be associated with a change in $b$ in the opposite direction. In case
of $0$ correlation, a change in $a$ is independent from $b$.

Autocorrelation describes the correlation of $a$ with itself at a different 
point in time. If $a$ is seasonal, then its autocorrelation will be seasonal
with the same season length. This property ensures that investigating 
autocorrelation leads to the same season length as analyzing the original 
observations, which is usually more challenging.

\subsection{Interpolating the Observations}

Autocorrelation should make it possible to observe seasonality.
Yet when applied on time series being preprocessed via low-pass filters one will experience a negative side-effect: 
Its characteristic subsequence's length $|Q_{A_x}|$ no longer matches the length original 
subsequence $|Q_x|$. Instead, its unscaled season length is $|Q_{A_x}| = k|Q_x|$  with $k \in \mathbb{N}$
and consequently a multiple of the original length, which makes finding
the correct season length more difficult.

To avoid this side-effect, it is possible to linearly interpolate the 
observations before applying a filter. This means that between every neighboring
observations, several intermediate observations are inserted. For example, the
sequence $y=\{0,2,1,2\}$ could be interpolated to 
$y'=\{\mathbf{0},1,\mathbf{2},1.5,\mathbf{1},1.5,\mathbf{2}\}$. The linear
interpolation $\psi(t)$ of two observations $x_1$ and $x_2$ can be written as
\begin{equation}
\psi(\Delta_t) = x_1 + \frac{x_2 - x_1}{t_{x_2} - t_{x_1}}(\Delta_t-t_{x_1})
\label{eq:interpolation}	
\end{equation}

where $t_{x_i}$ is the time at which $x_i$ was observed and $\Delta_t$ denotes
a chosen point in time between these two observations. By interpolating all 
neighboring observations, $x$ can be expanded to its interpolated sequence 
$x'$

Interpolating the observations is advantageous, since it usually lessens
the filters negative effect on the length of the characteristic subsequence.
Let $\chi$ be $x$ after interpolation, filtering and detrending. 
The unscaled season length of its autocorrelation $A_\chi$  matches 
the unscaled season length of the original 
observations $x$, which means $|Q_{A_\chi}| = |Q_x|$

\begin{figure}[!h]
	\centering
	\includegraphics[scale=0.45]{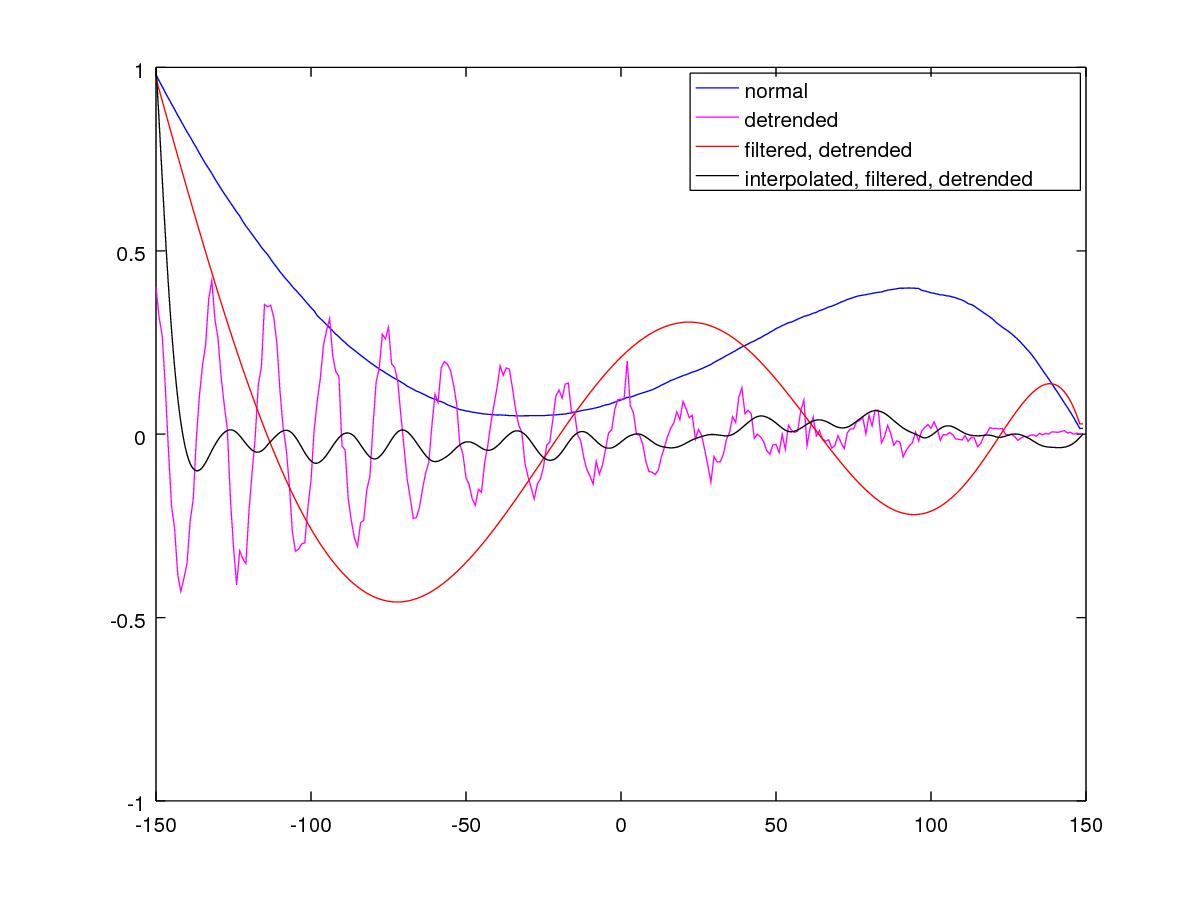}	
	\caption{Autocorrelation plot computed after three preprocessing steps - after all three steps have been applied, the autocorrelation will be further analyzed.}
\end{figure}

After interpolating, filtering and detrending the observations and calculating 
their autocorrelation, it is finally
possible to directly compute the time series' unscaled season length $s = 
|Q_{A_\chi}|$. 
This can be done by analyzing the zeros of $A_\chi$. The distance between two
zeros $A_\chi(\tau_1) = 0$ and $A_\chi(\tau_2) = 0$  with $\tau_1 < \tau_2$ 
is equal to half the unscaled season
length $s$. Therefore, $s$ can be computed with
\begin{equation}
s = |Q_{x}| = |Q_{A_\chi}| = 2(\tau_2 - \tau_1).
\label{eq:seasonlength}
\end{equation}

To calculate the true, scaled season length, one has to simply form the product 
$s\Delta_x$, which is negligible in practice.

\section{System}

\begin{figure}[!h]
	\centering
	\includegraphics[scale=0.45]{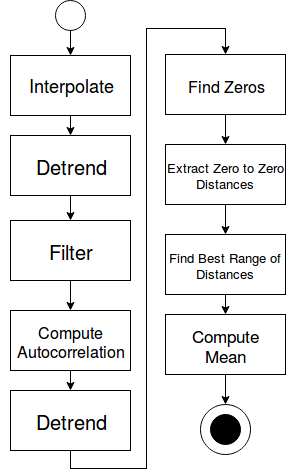}	
	\caption{The individual steps of the  season length detection system.}
	\label{fig:flowchart}
\end{figure}

While the above mentioned concepts work well in theory, there are several
challenges in practice which still need to be met:
\begin{itemize}
	\item Trends may change over time
	\item Noise may vary
	\item Machines have limited numeric precision
	\item Seasonality is not always perfect
	\item Presence of outliers
\end{itemize}
Dealing with these and other problems is imperative for robust parameter-free 
season length detection. An overview of the steps which our system takes to 
meet these challenges can be seen in Figure~\ref{fig:flowchart}.

Another crucial practical aspect is
runtime, since time series tend to contain thousands if not millions of 
observations. To be valuable in practice, it is desirable for the 
implementation to have a
worst-case computational complexity of at the most $\mathcal{O}(n\log n)$, with $n$ being
the number of observations in the time series.

\subsection{Approximating the Trend}

The seasonality of a time series typically revolves along a certain trend.
However, this trend may follow an arbitrary function, which makes removing
all trend in any time series impossible. Therefore, it is necessary to only
consider trends which are particularly common in practice.

Linear trends occur frequently in mathematical functions, but are not very 
common in practice. Therefore, it is insufficient to assume that all time series
will have a close to linear trend. However, as polynomials of higher order tend 
to over-fit the present data, linear trends can at least be compared with other
assumptions without over-fitting the data. Consequently, it is useful to begin
trend approximation by solving the 1st order polynomial linear regression.

Unlike linear tendencies, quadratic trends are fairly common. Many relations
in nature follow the square - for instance the relation between speed and 
acceleration of moving objects. Therefore it is advantageous to attempt 
modeling the data with a quadratic trend. To compare such a model with a linear
trend estimation one can compare the resulting mean square errors, which can
be achieved with
\begin{equation}
\ln(C(\theta_1) - C(\theta_2)) > k 
\label{eq:constant1}
\end{equation}

where $\theta_i$ are the parameters which minimize the mean square error 
$C(\theta)$ with a polynomial of $i$th degree. The logarithm is applied since
the difference tends to be very large. The constant $k$ determines the decision
boundary between linear and quadratic trend estimation. In the context of this
implementation it will be assumed that $k = e^2$. This value was chosen 
empirically rather than theoretically, which obviously will not be appropriate
in all cases. However, determining the correct value of this and a few other 
empirically chosen values for all cases would be beyond the scope of this paper.



\subsection{Choosing the Correct Filter}

A promising approach for detrending time series is high-pass 
filtering. An advantage of this approach is that it can effectively remove 
non-stationary trends from time series, which is difficult to achieve with 
linear regression. A disadvantage of high-pass filters is that they depend more
strongly on threshold values, which limits their value to that of the chosen 
models' constants.

There is a variety of ways to construct a low-pass filter, yet none of them
can create the ideal low-pass filter, which rejects all frequencies higher than
the cutoff frequency. It is only possible to approximate a close-to ideal 
filter, which removes most too high frequencies while changing the less frequent
signals as little as possible. This can be achieved by generating a Butterworth
low-pass filter, which can be obtained by choosing its order and cutoff 
frequency and then applying the corresponding algorithm.

However, choosing the correct order and cutoff frequency is problematic, as
the amount and amplitude of noise vary in time series. In the context of our
implementation, the order $\eta=2$ and cutoff-frequency $\omega=0.001\pi$ were 
chosen empirically in preliminary tests.

\subsection{Numeric Inaccuracies}
Machines only have a finite numeric precision.
Therefore, the theoretical method must be adjusted at several points.

Firstly,
there usually is no exact $0$ in the autocorrelation, rather the data moves from
positive to negative values or vice versa. Consequently, it is necessary to
either create a tolerance interval $\epsilon$ around $A_\chi(\tau)=0$, or to interpolate the data until
it is accurate enough. In case of our implementation, both strategies are applied.

Secondly, even after an almost ideal detrending, the autocorrelation will not
have a linear trend of $0$, but rather a tendency very close to $0$. To deal 
with this, a second linear regression is performed after the autocorrelation was
computed. The autocorrelation is then not searched for $A_\chi(\tau)=0$, but
rather $A_\chi(\tau)-X\theta=0$, where $X\theta$ is the linear trend of the 
$A_\chi$.

Thirdly, due to the consequential mentioned tolerance interval 
\begin{equation}
A_\chi(\tau)-X\theta \pm \epsilon=0	
\label{eq:toleranceInterval}
\end{equation}
there will be several solutions for $\tau$ which are in the
same tolerance interval. Therefore, it is necessary to discard all $\tau_1$ and
$\tau_2$ for which the following condition is true
\begin{equation}
|\tau_1 - \tau_2| \leq 1.
\label{eq:removeOnes}
\end{equation}

This is obvious, since no time series can have a season length shorter than two 
observations and the difference between two zeros corresponds to half the 
season length.

\subsection{Imperfect Seasonality}
Another problem that exists in practice is that seasonality is rarely 
perfect, as assumed in theory. Frequently, seasonality outliers or a 
systematic seasonal 
change occur, which invalidate the above formula $s=2(\tau_2 - \tau_1)$. To 
deal with these problems, it is possible to not only analyze $\tau_2$ and 
$\tau_1$, but rather any pair of adjacent solutions for Equation~\ref{eq:seasonlength}.

Let $\alpha=\{\alpha_1,\alpha_2,...,\alpha_m\}$ be all values for $\tau$ so
that Equation~\ref{eq:seasonlength} is fulfilled and let 
$\delta=\{\delta_1,...,\delta_{m-1}\} = \{\alpha_2-\alpha_1,\alpha_3-\alpha_2
,...,\alpha_m-\alpha_{m-1}\}$ be the distances between each pair of adjacent 
zeros in Equation~\ref{eq:seasonlength}. By removing all pairs that violate Equation~\ref{eq:removeOnes} from $\delta$ one
can compute all true distances between seasonal repetitions
$\delta'=\{\delta'_1,...,\delta'_l\}$, which are further sorted in ascending 
order.

These true distances $\delta'$ provide a better representation of the season 
length than an arbitrary pair of solutions $\tau_1$ and $\tau_2$. However, 
simply taking an average of these values does not suffice, since many time 
series contain seasonality outliers. Further, numeric imprecision may
cause the implementation to miss correct zeros, or any remaining noise might 
add incorrect zeros. Therefore, it is necessary to perform additional 
operations on $\delta'$.

To separate the correct values in $\delta'$ from the incorrect, it is 
necessary to assume that $\delta'$ contains a sufficiently large number of
correct values. Since seasonality is stationary in time series, the correctly
identified distances have a low variance in most cases, while erroneous 
values tend have a much higher sample variance.

When considering Equation~\ref{fig:delta}, it can be observed that the intervals 
$\delta'_{[3,8]}$ and $\delta'_{[9,10]}$ have a low variance when compared
to intervals including other values.
\begin{figure}[H]
	
\begin{equation}
 \delta' = [281,546,697,703,704,705,706,706,1411,1411,2823]
 \label{fig:delta}
\end{equation}
\caption*{A typical zero-to-zero distance vector}
\end{figure}
 This suggests that the correct 
zero-to-zero distance  is either $\frac{1}{2}s\approx703$ or
$\frac{1}{2}s\approx1411$. By further considering that 
$703 * 2 \approx 1411$, it is intuitive to assume that these intervals are
multiples of each other. Since Equation~\ref{eq:criterion2} must still be valid, it is safe to 
assume that a zero was missed between two zeros and thus caused the 
multiplied distance. 
Therefore, the correct zero-to-zero distance is very likely $703$.

While the above example suggests taking the low-variance interval with
the smallest mean, it is more reliable to rather search for the 
low-variance interval with the highest cardinality. The reason for this is the necessary assumption that 
a sufficiently large amount of the found zero-to-zero distances is almost correct,
as otherwise reliably finding the true season length would be impossible.

To apply the above reasoning in a generalized way, it is necessary to
reliably identify these low-variance intervals. This can be done by 
computing the quotients between the distances 
\begin{equation}
\gamma = 
\{\gamma_1,\gamma_2,...,\gamma_{l-1}\} = \{\frac{\delta'_2}{\delta'_1},
\frac{\delta'_3}{\delta'_2},...,
\frac{\delta'_l}{\delta'_{l-1}}\}.	
\label{eq:gamma}
\end{equation}

These quotients describe the rate at which the distances between zeros 
change. A series of low values for $\gamma$ implies a stable, low-variance
interval of distances, while high values suggest a jump between intervals.
These high and low values can be separated with
\begin{equation}
	\Gamma_i = 
	\begin{cases}
	1 & \text{if } i=1 \land |\gamma_1 - \gamma_2| \leq k\\
	i & \text{if } i=l-1\\
	i+1 & \text{if }|\gamma_i - \gamma_{i+1}| > k\\
	0 & \text{else}
	\end{cases}
	\label{eq:Gamma}
\end{equation}
where $k$ is a constant which was chosen empirically in preliminary tests with $0<k<1$ and 
$i={1,2,...,l-1}$.
This gives a series of integers where $0$ represents a low change of 
distances, and all other numbers the index of a high change. By then 
discarding all zeros with $\Gamma' = \Gamma\setminus\{0\}$, the longest
low-variance interval of distances can be found with
\begin{equation}
	i^* = \underset{i}{argmax}(\Gamma'_{i+1} - \Gamma'_{i})
	\label{eq:imax}
\end{equation}

\[
	a = \Gamma'_{i^*}
\]
\[
	b = \Gamma'_{i^* +1}
\]

where $a$ is the index of the first distance in the interval, and $b$ the
index of the last. If the resulting interval $\delta'_{[a,b]}$ is long enough, it can be advantageous do further 
discard an upper and lower percentile of the values within the interval. 
However, in practice, we found that this alters the result very little in our experiments.

With the longest low-variance interval, an approximation of the season 
length can be computed by taking the average of these distances with

\begin{equation}
	s = \frac{2}{b-a}
	\sum_{j=a}^{b}\delta'_j.
	\label{eq:result}
\end{equation}

\section{Evaluation}

\subsection{Data Set}

To evaluate the implementation, it is necessary to test it with several 
different time series. For this purpose, two distinct time series databases 
were gathered. The first database contained extensive variations of synthetic data, while the second database consisted of real financial and climate data.
The time series from these databases were tested in nine test runs, where our
implementation was tested against the existing algorithm in the R
forecast library. The test cases for runs 1 to 7 were taken from the synthetic 
database, while the test cases for runs 8 and 9 were taken from the real 
database and are briefly described below:

In the $Diverse$ test, both algorithms were confronted with 20 time series, 
which strongly vary in trend, seasonality, noise and length. However, all
these time series had a consistent season length, which does not vary
within the series. This first testing was meant to compare the general
applicability of the algorithm, which is achieved by challenging them with
very different problems, which yet have a distinct solution.

In the $Complex$ test, the 20 test cases included many numeric outliers, 
largely changing season amplitude, season outliers and also varying
noise. This had the purpose to assess the algorithms' error tolerance.

The $Ambiguous$ test included 20 examples with more than one correct solution. 
This aimed at observing the choices made by the algorithms and thus identifying 
their tendencies or preferences.

The $Variations$ test featured 4 time series, which were each presented in 
5 variations. This was meant to assess the algorithms' consistency.

The $Noise$ test started with a simple time series without any noise, which
was then tested repeatedly with increasing amounts of noise. The goal of this 
run was to compare the noise resilience of the two algorithms.

The $Length$ test also consisted of a single, simple time series, which was 
presented with varying season length. The purpose of this run was to test both 
algorithms in different result domains.

The $NoSeason$ test featured 10 examples with no seasonality. This had the 
purpose to explore the algorithms' capability to distinguish between seasonal 
and non-seasonal time series.

The $Econonmy$ test consisted of 20 time series of seasonal economic data. 
These financial data were taken primarily from sectors which display seasonality, such as tourisms and restaurants.
Three of these test cases contained both annual and quarterly seasonality, and 
several others only showed a very slight manifestation of seasonality.

The $Climate$ test contained 20 time series of seasonal climate data about 
temperature, precipitation, sun hours or storm counts. 

\subsection{Results}

During testing, the R library algorithm which uses spectral density estimation was referred to as \textit{Spectral} and our system using autocorrelation as \textit{Autocorr}. For each test, the 
season length suggested by the algorithms was considered correct if it was
within an error margin of $\pm20\%$ of the reference value.

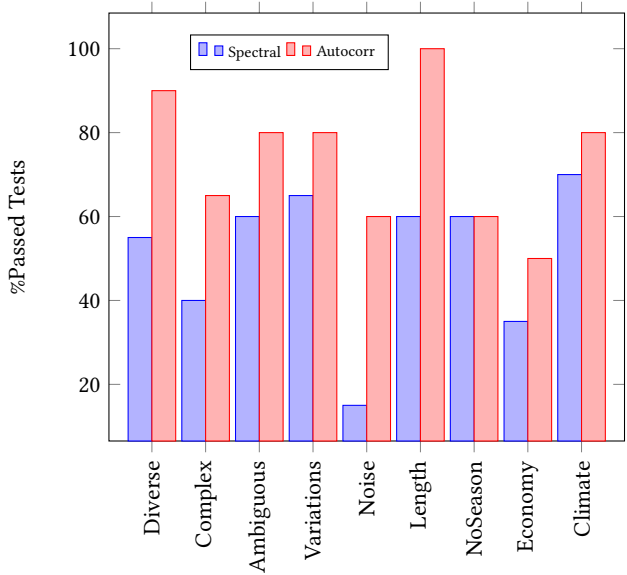
\begin{figure}[!h]
\centering
\begin{tikzpicture}[scale=1.0] \begin{axis}[x tick label style={rotate=90,anchor=east}, ylabel=\%Passed Tests, enlargelimits=0.1, legend style={at={(0.35,0.95),font=\scriptsize}, anchor=north,legend columns=-1}, ybar=0pt,
bar width=9pt, point meta=y, 
xtick=data,
xticklabels={Diverse,Complex,Ambiguous,Variations,Noise,Length,NoSeason,Economy,Climate}
] \addplot coordinates {(1,55) (2,40) (3,60) (4,65) (5,15) (6,60) (7,60) (8,35) (9,70)}; \addplot coordinates {(1,90) (2,65) (3,80) (4,80) (5,60) (6,100) (7,60) (8,50) (9,80)}; \legend{Spectral,Autocorr} \end{axis} \end{tikzpicture}
\caption{Test results of algorithms \textit{Spectral} and \textit{Autocorr} for all 9 test runs.}

\label{fig:results}	
\end{figure}

The accumulated test results are depicted in Table~\ref{tab:results}, while the results of the individual test runs can be seen in Figure~\ref{fig:results}.
In the first 6 tests algorithm $Autocorr$ performed better than algorithm $Spectral$. The most
significant difference was in the noise test, were $Spectral$ passed 3 tests 
and $Autocorr$ 12 tests. 
The $NoSeason$ test, which only consisted of time series without seasonality,
was the only test run were both algorithms passed the same number of tests. In 
all other test runs, including $Economy$ and $Climate$, $Autocorr$ passed more
tests than $Spectral$.
\begin{table}[h!]

\begin{tabular}{c|c |c c}
Database&\#Test Cases&Spectral&Autocorr \\
\hline Synthetic&125&62&96\\
Real&40&21&26 \\
\hline$\sum$&165&83&122\\

\end{tabular}
\caption{Numeric Test Results with the number of correctly identified season lengths for both algorithms.}
\label{tab:results}
\end{table}

\begin{figure}
	\includegraphics[scale=0.45]{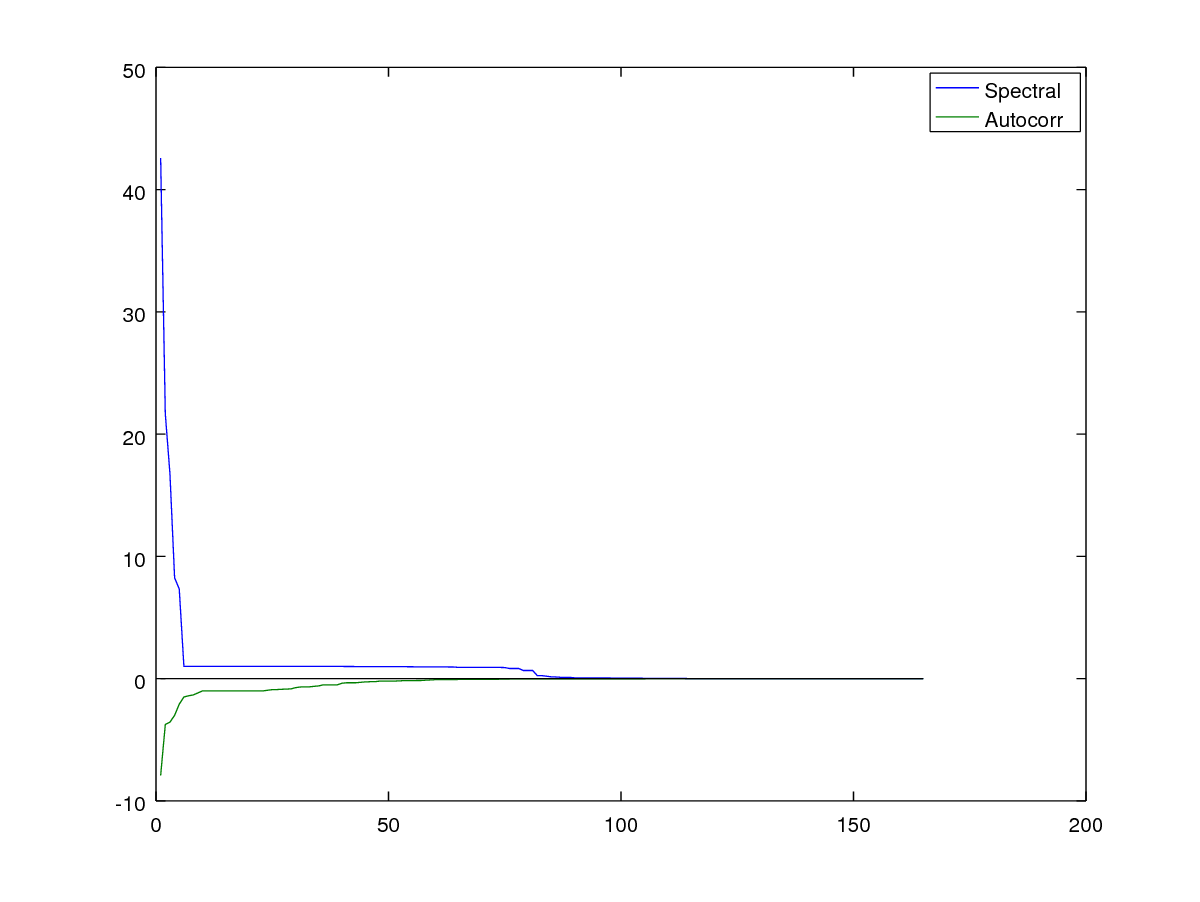}
	
	\caption{Relative detection error in all test cases - \textit{Autocorr}'s curve was negated to contrast the deviations}
	\label{fig:errors}
\end{figure}

Figure~\ref{fig:errors} displays the relative difference between the detected
season length and the reference value for all test cases. The area between the
curves and zero represents the accumulated test error of the algorithms. 
\textit{Spectral} had an accumulated error of 172\%, while \textit{Autocorr}'s 
relative errors add up to 57\%.

\section{Discussion}

The purpose of this paper was to present an algorithm developed for identifying time 
series' season length which is sufficiently reliable to be used in real-world applications. 
The conducted tests tried to cover many different aspects of time series behavior and our system compares favorably against the reference algorithm. 

However, the overall results alone do not necessarily suggest that the methods
applied in $Autocorr$ are preferable for identifying season length in all settings. Both 
implementations still rely on empirically chosen constants, which may have an influence on the outcome. 
For example, had the constant in $Autocorr$ for distinguishing 
linear and quadratic trends been chosen slightly higher, then several time
series with quadratic trends may have been misclassified. This argument can
be made for every empiric constant in both implementations.

The $Noise$ test, which was a repeated test of the same time series while 
increasing the noise component from test to test, has revealed a noise 
susceptibility of $Spectral$. This is likely an inherent property of working with
spectral density estimates, as noise tends to hide the relevant frequencies.
This is also supported by the $Variations$ test, where $Spectral$ failed to disregard
outliers induced on otherwise unchanged time series.

Another noteworthy observation is that $Spectral$ frequently misclassified season 
lengths longer than $100\Delta_x$. In fact, $Spectral$ never suggested a season
length longer than $998\Delta_x$, although there were 18 cases where this 
would have been necessary. This suggests that the implementation of $Spectral$ may not be the optimal choice for
investigating seasonality in large time series with very long season lengths.
Additional evidence for this assumption is provided by the $Length$ test, which was
designed to test this particular aspect of seasonality.

A dependency that greatly affects both algorithms is that they both rely on an adequate
removal of trend from the time series. Algorithm $Autocorr$ always returned an 
erroneous result if it failed to correctly identify the trend. Moreover, 
neither $Spectral$ nor $Autocorr$ were capable of detrending any non-linear or non-square 
trend. Correctly removing trends from time series is an active research field, fostering the hope that this issue will be addressed.

There is also a single case of a false positive in $Autocorr$ that is not directly apparent. While it did correctly
analyze the season length of test case 5 in the the $NoSeason$ test, it was only 
coincidental that the result is correct. The reason for this is that
$Autocorr$ perfectly removed all trend from this purely quadratic time series 
and then considered almost every not interpolated point in the 
autocorrelation as potential zero. Computing the mean distance between
these obviously yields an average of 1, which is then discarded due to 
Equation~\ref{eq:removeOnes}. However, with an only slightly altered time series, the 
average distance could be rounded up to 2, which would not be discarded, yet 
still incorrect for a purely quadratic time series.


The most expensive operation in both implementations is computing the 
matrix-inverse required for detrending the data, which has a worst-case 
computational complexity of $\mathcal{O}(n^{2.3727})$. This is far above the desirable
computational complexity of $\mathcal{O}(n\log n)$, as for an exemplar time series with
$1000$ observations, in the ideal case $k * 10^3$ operations would be required,
whereas both implementations need in the worst case $k * 10^6$ operations, which
is a thousand times higher.

\section{Conclusion}

Detecting season length in time series without human assistance is challenging. The method presented in this paper attempts to complete
this task by interpolating, filtering and detrending a time series and then
analyzing the distances between zeros in its autocorrelation function. The implementation
of this concept is still leaves room for improvement, as it still relies on several empirically 
chosen constants. However, the results demonstrated sufficient robustness in our evaluation.

Future work concerning automated season length detection might attempt to 
eliminate the algorithm's constant-dependency by inferring them from the data.
Further, the detrending can likely be improved by including high-pass filtering
into the procedure. Another interesting concept would be to develop a season 
length detection algorithm based on a machine learning method like a neural 
network.


To advance automated seasonality and periodicity mining in general, it is 
important to develop automated procedures for a dynamic computation of 
otherwise static constants in contemporary algorithms. Achieving this is an
interesting challenge for the future.

\bibliographystyle{ACM-Reference-Format}
\bibliography{references-biblatex.bib} 

\end{document}